# Choosing Among Interpretations of Probability

**Henry E. Kyburg, Jr.**
Philosophy and Computer Science
University of Rochester
Rochester, NY 14627, USA

**Choh Man Teng**
Computer Science and Engineering
University of New South Wales
Sydney, NSW 2052, Australia

## Abstract

There is available an ever-increasing variety of procedures for managing uncertainty. These methods are discussed in the literature of artificial intelligence, as well as in the literature of philosophy of science. Heretofore these methods have been evaluated by intuition, discussion, and the general philosophical method of argument and counterexample. Almost any method of uncertainty management will have the property that in the *long run* it will deliver numbers approaching the relative frequency of the kinds of events at issue. To find a measure that will provide a meaningful evaluation of these treatments of uncertainty, we must look, not at the long run, but at the short or intermediate run. Our project attempts to develop such a measure in terms of short or intermediate length performance. We represent the effects of practical choices by the outcomes of bets offered to agents characterized by two uncertainty management approaches: the subjective Bayesian approach and the Classical confidence interval approach. Experimental evaluation suggests that the confidence interval approach can outperform the subjective approach in the relatively short run.

## 1 Introduction

Probability is the very guide of life, said Bishop Butler, and many agree. But what that comes to depends on what you mean by "probability". In the past there have been several clearcut interpretations of probability to choose from: The subjective Bayesian view, according to which a probability represents an individual's actual or idealised degree of belief [Savage, 1954], the logical view, according to which the probability of a statement, relative to a body of evidence, represents something like partial entailment, and is a relation between the *evidence* an individual has and that statement [Carnap, 1950], and the empirical view which includes as variants finite frequency [Russell, 1901], limiting frequency [Mises, 1957], and measure-theoretic [Cramér, 1951] views according to which a probability statement is a claim about the world of the same general character as a universal generalization or a law.

In recent years this neat trichotomy has become less sharp. Many writers seem to have decided that vagueness is the better part of valor: One may assign probabilities to statements without specifying where the assignment comes from or what it comes to — whether the source is logical measure or intuitive belief, whether it is intended to be representational or normative.

From the point of view of the pure logician, this makes perfect sense: just as logic is not (in general) concerned to tell us what statements are true, but rather to spell out relations among truth assignments, so probabilistic logic need not be concerned with the probability assigned to any particular statement, but only with the relations of probabilities assigned to statements in related groups — for example, an algebra of statements. There is thus a reasonable precedent for leaving the problem of interpreting probability to someone else.

## 2 There is an Issue

Despite the atmosphere of tolerance and good will (or perhaps fatigue) that currently surrounds questions of the interpretation of probability, there are important issues to be discussed, and perhaps even resolved. The issues are important because probability (or some measure of uncertainty) is central in decision making under the only circumstances we have access to.

There is no extrasystematic way of determining whether the probability assigned to a statement ("Pe-ter's bike will not be stolen," "Susan's toss will land tails") is "correct" or not. By this we mean that the truth or falsity of the statement in question ("The bike is stolen," "The coin lands tails") cannot confirm or refute the probability we have assigned to it. The improbable happens (my bridge hand may consist of thirteen cards of the same suit) and what is almost certain to occur may not occur (a large fair sample may turn out to be totally misleading). Of course *within* some system we can confirm or refute the assignment of probabilities: the subjective Bayesian can examine his (or your) sincere propensity to bet; the logical theorist can compute the measures of the relevant models. However, any statement is either true in our one world, or false in it, and that is the end of the matter. The truth or falsity of the single statement gives us no handle on its probability, no way of evaluating the probability claim.

A natural response is to say that while the truth or falsity of a single statement says nothing about its probability, yet in the long run the relative frequency of truth in a class of statements of probability $p$ ought to come close to $p$. But the connection between long run relative frequency of truth and probability is itself a probabilistic connection: we should expect that sometimes the highly probable turns out to be false. The only way to make this connection tight is to adopt a frequency interpretation of probability, and that breaks the very connection we need: probability no longer applies to the single case.

What we are looking for is a way to evaluate the different approaches to uncertainty that is more convincing than calling on conflicting intuitions. These would include not only different interpretations of probability, but also so-called non-probabilistic approaches such as Dempster/Shafer belief functions [Shafer, 1976] and possibility theory [Dubois and Prade, 1992].

Here we will concentrate on comparing the subjective Bayesian approach and the classical confidence interval approach to probability.

Confidence methods bridge the gap between frequencies and probabilities of particular cases by allowing us to accept (or to "fail to reject"), at a certain level of confidence, statistical hypotheses that impose interval constraints on long run frequencies.[1] The agent accepts that the long run relative frequency lies between an upper and a lower bound. The confidence level represents practical certainty: one minus the chance of being wrong we are just willing to tolerate.

The subjective Bayesian approach eschews acceptance, yielding only probability distributions over hypotheses. But when it comes to specific events, it yields point-valued probabilities, and thus the principle of maximizing expectation is always applicable.

## 3 Comparing the Performance

A natural idea is to consider, not one statement, but many. In the simplest case, we might look at a sequence of *trials*, to which alternative conceptions of probability may be applied. In the long run we have no test that compares the sensible (non-extreme) Bayesian to the probabilist who bases his probabilities on confidence limits at a given level of confidence. For a fixed confidence level, as we consider more trials in the sequence, an ever narrower confidence interval can, at that fixed level of confidence, be taken to characterize the sequence. The non-extreme subjective Bayesian will have beliefs that converge to the same long run relative frequency approached by the progressively narrower confidence intervals.

Thus, "in the long run" there is no difference between the two approaches. Presumably this is also true of any sensible alternative view — a view of probability that did *not* lead to convergence towards an observed frequency in a sequence of this sort in the long run would surely be strange.

This is however not true of the short run. What we examine is whether the frequency-based probabilist is *better off* than the subjective Bayesian or vice versa in the short run.[2] Although we know that they will come close to agreement in the long run, there may be a statistically significant difference in favor of one or the other in the short run.

One way to get at this is to have proponents of the two views bet with each other on a sequence of trials of an event with an unknown probability of success.[3] Each would represent and update its own beliefs with respect to the interpretation of probability it adheres to, and place bets on the success of the trials accordingly. The profit or loss from the bets on a (not very long) sequence of trials can then serve as an indicator of how well a particular approach works.

## 4 Rules of the Game

In this section we describe the set up of the game we used to evaluate the performance of the Bayesian and confidence interval approaches. Let us call the two players *Bayes*, an adherent of subjective Bayesianism, and *Conf*, an adherent of classical confidence methods.

---

[1]More general procedures, based on imprecise probabilities, are discussed by Peter Walley [Walley, 1995].

[2]The confidence theorist does not need a *very* long run to get going: at a confidence of 0.75 he can be confident that in a sample of two the sample ratio will differ by no more than 1/3 from the unknown parameter $p$.

[3]This was first suggested by Murtezaoglu in his dissertation [Murtezaoglu, 1998].

| Action | Outcome heads | tails |
|---|---|---|
| buy | $1-t$ | $-t$ |
| sell | $-(1-t)$ | $t$ |
| hold | 0 | 0 |

Table 1: Payoffs of a trial with ticket price $t

The game consists of a sequence of tosses of a (possibly biased) coin, whose chance of turning up heads is $p$. The two agents bet on the outcomes of the tosses according to their own estimates of the probability of heads. Before each toss, a random price $t between $0 and $1 is posted as the price of a ticket that will return $1 if the next toss is heads, and nothing otherwise. We make a market in those tickets: that is, we are willing to buy or sell any number of tickets at price $t. A new randomly generated price is adopted for each new trial.

Given the ticket price $t for a trial, *Bayes* and *Conf* each has three options: buy a ticket at price $t, sell a ticket at price $(1-t)$, or hold (decline to participate in this trial). The payoffs are given in Table 1.

### 4.1 Belief Representation

The two players represent their beliefs differently. *Bayes* represents its belief by a beta distribution, which is a commonly assumed subjectivist distribution and one that is easy to update. *Conf* represents its belief by a confidence interval at confidence level $1-\alpha$. At the start of the game, neither of the agents has any information about the coin. *Bayes* starts with a prior distribution beta$(a,b)$. (A flat prior distribution in which every value of $p$ is equally likely is given by beta$(1,1)$). *Conf* on the other hand represents total ignorance by the widest interval possible; that is, $[0,1]$. Its initial state of belief may also be thought of as accepting as possible the whole *set* of Bernoulli distributions.

As more evidence comes in, in the form of outcomes of previous coin tosses in the sequence of trials, the two agents each update their beliefs in heads according to their own scheme. *Bayes* changes its belief to beta$(a+h, b+\bar{h})$, where $h$ and $\bar{h}$ are the numbers of heads and tails observed so far. *Conf* derives the interval at confidence $1-\alpha$ based on $h$ and $t$, such that $1-\alpha$ of the time the real probability $p$ falls within this interval. It rejects the Bernoulli distributions with values of $p$ falling outside this interval.

### 4.2 Betting Strategy

Given a ticket price of $t, and a mean $x$ of the current beta distribution, *Bayes* buys a ticket if $t \leq x$, and sells a ticket otherwise. *Conf* with confidence interval $[l, u]$ buys a ticket if $t < l$, sells a ticket if $t > u$, and declines to bet if $t$ falls within the bounds of the interval.

It is not clear what the fairest procedure is. *Bayes* always has an exact probability, and so can always choose a bet with the maximum expectation. On the other hand, *Conf* is almost sure to start the sequence by declining all bets, until it has accumulated some knowledge of the distribution. We have chosen to take account of this by allowing each player to bet no more than $m$ times on the $n$ tosses.

For each trial, each has to decide whether to use a token or not on the spot, and once we have moved on to the next toss, they cannot change their minds about decisions made for the previous tosses and go back to bet on a previously declined trial. This eliminates strategies that require global planning or lookahead, such as choosing the $m$ trials whose ticket prices are the most extreme among the $n$ trials.

This presents an added complication for *Bayes*: *Bayes* must decide, on each toss, whether to bet on that toss or to save its bet until later when it (presumably) has a "better" posterior probability on which to base its bet. *Conf* follows a simpler algorithm, but faces the prospect of not being able to use all of its $m$ bets. To simplify matters we have supposed that *Bayes* will bet on the last $m$ of the $n$ opportunities to bet, and that *Conf* will simply bet whenever it reasonably can. To compensate for these difficulties we have varied the number of bets $m$ as a fraction of the number of opportunities $n$.

## 5 Preliminary Evaluation

We set up the game using the following parameters. 100 runs were performed with each combination of parameter values.

**Chance of heads $p$:** $\{0.1, 0.3, 0.5, 0.7, 0.9\}$

**Length of the binomial sequence of trials $n$:** $\{3, 5, 10, 20, 30, 50\}$

**Number of tokens $m$:** $\{0.1, 0.3, 0.5, 0.7, 1.0\}$ of the number of trials $n$

**Confidence level $1-\alpha$:** $\{0.5, 0.6, 0.7, 0.8, 0.9, 0.95, 0.99\}$ (or $\alpha \in \{0.5, 0.4, 0.3, 0.2, 0.1, 0.05, 0.01\}$)

**Priors (beta($a, b$)):**

$\{(1,1), (1, k+1), (k+1, 1), (k+1, k+1)\}$, for $k \in \{0.1, 0.3, 0.5, 0.7, 1.0\}$ of the number of trials $n$

Confidence levels are traditionally chosen in the range 0.90 to 0.99. On the other hand, they are perfectly well defined for any parameter greater than 0.50. Low confidence may not seem like much, but if it represents "more likely than not", and is based on objective facts about the world, it becomes more interesting. Note also that confidence methods can yield substantive conclusions from even small samples.

Note that under the rules of the game, the long run relative frequency of heads $p$ is not a global parameter we set, but rather one the agents are trying to get a good estimate of. We have included it as a parameter in the evaluation however, to see how the performance changes with respect to the target relative frequency. Results are given broken down by individual $p$ values as well as averaged over all settings of $p$.

The net profits per allowed bet averaged over 100 runs are summarized in Tables 2 to 5. Due to the large number of combinations of parameters, we present only a cross section of results in each table with some of the parameters fixed, to give an idea of how the net profit varies with a particular parameter. Table 2 gives the results obtained from different numbers of trials $n$. Table 3 reports on the effect of changes in the number of tokens $m$. Tables 4 and 5 show respectively how the net profit of *Conf* varied with the confidence level $1 - \alpha$, and how the net profit of *Bayes* varied with the priors given by the beta distribution beta($a, b$).

For comparison, we also included the results obtained using the sample proportion (*Sample*) as the estimate. *Sample* buys a ticket at price $\$t$ if $t \leq h/(h+\bar{h})$, where $h$ and $\bar{h}$ are the numbers of heads and tails observed so far; otherwise it sells a ticket. Just as for the other two players, it is constrained by the maximum number of bets $m$ it can place. The betting strategy of *Sample* is the same as that of *Bayes*. It bets on the last $m$ trials of the sequence, as it should by then have a better estimate of the probability of heads.

Note that *Sample* and *Bayes* always place the maximum allowable number of bets, while *Conf* might not. The profit however was averaged over the *maximum* allowable number ($m$), not the *actual* number of bets placed.[4]

---

[4] If we consider the profit per *actual* number of bets placed, *Conf* scored substantially better than the other two agents.

| $p$ | $n$ | *Sample* | *Bayes* | *Conf* |
|---|---|---|---|---|
| 0.1 | 3 | 0.2664 | 0.2981 | 0.0345 |
|  | 5 | 0.3478 | 0.3717 | 0.1711 |
|  | 10 | 0.4021 | 0.4001 | 0.3433 |
|  | 20 | 0.3952 | 0.3939 | 0.4904 |
|  | 30 | 0.4092 | 0.4090 | 0.5491 |
|  | 50 | 0.3980 | 0.4002 | 0.5395 |
| 0.3 | 3 | 0.1102 | 0.2549 | 0.0470 |
|  | 5 | 0.1925 | 0.2401 | 0.0671 |
|  | 10 | 0.2635 | 0.2932 | 0.1920 |
|  | 20 | 0.3005 | 0.2913 | 0.3320 |
|  | 30 | 0.2729 | 0.2742 | 0.3633 |
|  | 50 | 0.2888 | 0.2923 | 0.3916 |
| 0.5 | 3 | 0.0799 | 0.2325 | 0.0698 |
|  | 5 | 0.1595 | 0.2024 | 0.1147 |
|  | 10 | 0.2317 | 0.2536 | 0.1759 |
|  | 20 | 0.2574 | 0.2538 | 0.2653 |
|  | 30 | 0.2352 | 0.2377 | 0.3017 |
|  | 50 | 0.2491 | 0.2491 | 0.3390 |
| 0.7 | 3 | 0.1985 | 0.2991 | 0.0744 |
|  | 5 | 0.2889 | 0.2916 | 0.2172 |
|  | 10 | 0.2442 | 0.2658 | 0.2680 |
|  | 20 | 0.2447 | 0.2529 | 0.3355 |
|  | 30 | 0.2781 | 0.2757 | 0.4002 |
|  | 50 | 0.2961 | 0.2962 | 0.4038 |
| 0.9 | 3 | 0.3831 | 0.3746 | 0.1103 |
|  | 5 | 0.4287 | 0.4161 | 0.3588 |
|  | 10 | 0.3787 | 0.3610 | 0.4636 |
|  | 20 | 0.4017 | 0.4016 | 0.5437 |
|  | 30 | 0.3966 | 0.3909 | 0.5533 |
|  | 50 | 0.4121 | 0.4108 | 0.5467 |
| Overall | 3 | 0.2076 | 0.2918 | 0.0672 |
|  | 5 | 0.2835 | 0.3044 | 0.1858 |
|  | 10 | 0.3040 | 0.3147 | 0.2886 |
|  | 20 | 0.3199 | 0.3187 | 0.3934 |
|  | 30 | 0.3184 | 0.3175 | 0.4335 |
|  | 50 | 0.3288 | 0.3297 | 0.4441 |

Table 2: Net profit per allowed bet, varying the number of trials $n$ ($m = 0.5n$, $\alpha = 0.1$, priors=beta(1, 1)).

## 6 Discussion

Let us examine the effects of varying the parameters of the experiments. First note that the net profits were positive in all cases, so all three methods of updating and betting have the right idea. The question is *how* right they are.

The net profit was higher in almost all settings when the chance $p$ of heads of the coin was more towards the two ends of the spectrum (0 and 1). The midpoint $p = 0.5$ gave the lowest yields. But of course the agents are not in control of this parameter. In fact, they do not even know its value; otherwise they would not be applying various methods to update their beliefs about $p$. The entries under "Overall" in Tables 2 to 5 show the net profits per allowed bet averaged over all settings of $p$, giving an indication of the overall performance of the agents across a wide range of chance of heads.

| p | m | Sample | Bayes | Conf |
|---|---|--------|-------|------|
| 0.1 | 2 | 0.3689 | 0.3795 | 0.6567 |
|  | 6 | 0.4021 | 0.4055 | 0.5973 |
|  | 10 | 0.3952 | 0.3939 | 0.4904 |
|  | 14 | 0.3961 | 0.3931 | 0.3623 |
|  | 20 | 0.3762 | 0.3743 | 0.2536 |
| 0.3 | 2 | 0.2995 | 0.2995 | 0.5003 |
|  | 6 | 0.3205 | 0.3123 | 0.4344 |
|  | 10 | 0.3005 | 0.2913 | 0.3320 |
|  | 14 | 0.2928 | 0.2886 | 0.2381 |
|  | 20 | 0.2649 | 0.2751 | 0.1667 |
| 0.5 | 2 | 0.2221 | 0.2069 | 0.4082 |
|  | 6 | 0.2565 | 0.2512 | 0.3509 |
|  | 10 | 0.2574 | 0.2538 | 0.2653 |
|  | 14 | 0.2409 | 0.2388 | 0.1919 |
|  | 20 | 0.2143 | 0.2325 | 0.1343 |
| 0.7 | 2 | 0.2281 | 0.2412 | 0.4473 |
|  | 6 | 0.2463 | 0.2543 | 0.4069 |
|  | 10 | 0.2447 | 0.2529 | 0.3355 |
|  | 14 | 0.2461 | 0.2545 | 0.2489 |
|  | 20 | 0.2375 | 0.2553 | 0.1741 |
| 0.9 | 2 | 0.4396 | 0.4316 | 0.6558 |
|  | 6 | 0.3992 | 0.4021 | 0.6081 |
|  | 10 | 0.4017 | 0.4016 | 0.5437 |
|  | 14 | 0.3989 | 0.3963 | 0.4371 |
|  | 20 | 0.3864 | 0.3831 | 0.3067 |
| Overall | 2 | 0.3116 | 0.3117 | 0.5337 |
|  | 6 | 0.3249 | 0.3251 | 0.4795 |
|  | 10 | 0.3199 | 0.3187 | 0.3934 |
|  | 14 | 0.3150 | 0.3143 | 0.2957 |
|  | 20 | 0.2959 | 0.3041 | 0.2071 |

Table 3: Net profit per allowed bet, varying the number of tokens $m$ ($n = 20$, $\alpha = 0.1$, priors=beta(1, 1)).

In the very short run ($n = 3$, Table 2), the net profits were low for all three players, but they increased with the length of the run $n$, very quickly at the beginning and then slower (and possibly flattened out) when $n$ became large. *Conf* started out much lower than *Bayes* and *Sample*, but overtook them both as $n$ was increased to the 10-20 range. We can expect that *Conf* does not bet much at the beginning, when its confidence interval is close to [0, 1]. An initial (not very long) sequence is needed to successively refine its interval estimate before bets can be placed effectively.

| p | $1 - \alpha$ | Sample | Bayes | Conf |
|---|---|--------|-------|------|
| 0.1 | 0.5 | 0.3952 | 0.3939 | 0.5095 |
|  | 0.6 |  |  | 0.5114 |
|  | 0.7 |  |  | 0.5074 |
|  | 0.8 |  |  | 0.5093 |
|  | 0.9 |  |  | 0.4904 |
|  | 0.95 |  |  | 0.4692 |
|  | 0.99 |  |  | 0.4039 |
| 0.3 | 0.5 | 0.3005 | 0.2913 | 0.3509 |
|  | 0.6 |  |  | 0.3545 |
|  | 0.7 |  |  | 0.3608 |
|  | 0.8 |  |  | 0.3559 |
|  | 0.9 |  |  | 0.3320 |
|  | 0.95 |  |  | 0.3048 |
|  | 0.99 |  |  | 0.2457 |
| 0.5 | 0.5 | 0.2574 | 0.2538 | 0.3027 |
|  | 0.6 |  |  | 0.3016 |
|  | 0.7 |  |  | 0.3016 |
|  | 0.8 |  |  | 0.2923 |
|  | 0.9 |  |  | 0.2653 |
|  | 0.95 |  |  | 0.2359 |
|  | 0.99 |  |  | 0.1875 |
| 0.7 | 0.5 | 0.2447 | 0.2529 | 0.3521 |
|  | 0.6 |  |  | 0.3517 |
|  | 0.7 |  |  | 0.3480 |
|  | 0.8 |  |  | 0.3474 |
|  | 0.9 |  |  | 0.3355 |
|  | 0.95 |  |  | 0.3056 |
|  | 0.99 |  |  | 0.2702 |
| 0.9 | 0.5 | 0.4017 | 0.4016 | 0.5252 |
|  | 0.6 |  |  | 0.5386 |
|  | 0.7 |  |  | 0.5455 |
|  | 0.8 |  |  | 0.5442 |
|  | 0.9 |  |  | 0.5437 |
|  | 0.95 |  |  | 0.5322 |
|  | 0.99 |  |  | 0.4965 |
| Overall | 0.5 | 0.3199 | 0.3187 | 0.4081 |
|  | 0.6 |  |  | 0.4116 |
|  | 0.7 |  |  | 0.4127 |
|  | 0.8 |  |  | 0.4098 |
|  | 0.9 |  |  | 0.3934 |
|  | 0.95 |  |  | 0.3695 |
|  | 0.99 |  |  | 0.3208 |

Table 4: Net profit per allowed bet, varying the level of confidence $1 - \alpha$ ($n = 20$, $m = 10$, priors=beta(1, 1)). Note that the results for *Sample* and *Bayes* do not vary with the level of confidence, and duplicate entries have been omitted.

| $p$ | $(a,b)$ | Sample | Bayes | Conf |
|---|---|---|---|---|
| 0.1 | (1,1) | 0.3952 | 0.3939 | 0.4904 |
|  | (11,1) |  | 0.2475 |  |
|  | (1,11) |  | 0.4104 |  |
|  | (11,11) |  | 0.3422 |  |
| 0.3 | (1,1) | 0.3005 | 0.2913 | 0.3320 |
|  | (11,1) |  | 0.2139 |  |
|  | (1,11) |  | 0.3034 |  |
|  | (11,11) |  | 0.2928 |  |
| 0.5 | (1,1) | 0.2574 | 0.2538 | 0.2653 |
|  | (11,1) |  | 0.2035 |  |
|  | (1,11) |  | 0.2197 |  |
|  | (11,11) |  | 0.2544 |  |
| 0.7 | (1,1) | 0.2447 | 0.2529 | 0.3355 |
|  | (11,1) |  | 0.2471 |  |
|  | (1,11) |  | 0.2106 |  |
|  | (11,11) |  | 0.2654 |  |
| 0.9 | (1,1) | 0.4017 | 0.4016 | 0.5437 |
|  | (11,1) |  | 0.4055 |  |
|  | (1,11) |  | 0.2663 |  |
|  | (11,11) |  | 0.3610 |  |
| Overall | (1,1) | 0.3199 | 0.3187 | 0.3934 |
|  | (11,1) |  | 0.2635 |  |
|  | (1,11) |  | 0.2821 |  |
|  | (11,11) |  | 0.3032 |  |

Table 5: Net profit per allowed bet, varying the priors given by the beta distribution beta$(a,b)$ ($n = 20$, $m = 10$, $\alpha = 0.1$). Note that the results for Sample and Conf do not vary with the beta distribution, and duplicate entries have been omitted.

Now consider the effect of varying the limit $m$, the maximum number of bets placed (Table 3). Sample and Bayes were not affected very much by $m$, but Conf's performance decreased drastically as $m$ increased. For Conf, not every trial is safe to bet on, and thus at higher $m$'s, it might not encounter enough "good" trials it is willing to place a bet on. However, since each bet it undertakes is likely to be of good value, the average return per bet would be higher, giving it an advantage over the other two players when $m$ is small.

The other two parameters we considered do not apply to every agent. The level of confidence $1 - \alpha$ is specific to Conf, and has no effect on Sample and Bayes. From Table 4, we see that the net profit for Conf decreased as the level of confidence increased. This effect again can be attributed to the number of bets placed. With respect to the same sequence of observations, as the level of confidence increases, the confidence interval obtained based on the sequence becomes wider. Thus the higher the level of confidence, the fewer ticket prices Conf would find attractive. It appeared that the lower confidence levels allowed for a better balance between the numbers of bets placed and declined.

The last parameter, the priors of the beta distribution beta$(a,b)$, affects only Bayes. It characterizes Bayes' initial bias on the probability of heads before it encounters any evidence. We only showed the results for four different priors in Table 5: $(1,1)$ is a flat distribution; $(11,1)$ favors heads; $(1,11)$ favors tails; and $(11,11)$ favors an equal distribution of heads and tails. The priors increased the profit of Bayes when they closely matched the real chance of heads $p$. For example, at $p = 0.1$, $(1,11)$ performed the best and $(11,1)$ the worst, while at $p = 0.9$, the reverse held. At $p = 0.5$, both $(1,1)$ and $(11,11)$ performed well, with the latter being slightly better.

Averaged over all settings of the chance of heads $p$, the uniform prior $(1,1)$ performed better than the other three priors. This supports the principle of indifference: it is best *not* to assume a biased prior in the absence of good reasons. The gain from a matching prior is much less than the loss incurred when there is a mismatch. It is also interesting that in every case, including the ones in which the priors matched the real chance of heads, Conf performed better than both Bayes and Sample.

## 7    More Thoughts

Speaking generally, Sample and Bayes had comparable performance, while Conf appeared to obtain higher returns in quite a few cases. Although Conf bet less often, on average the yield per actual bet was higher. However, the differences in performance could also partially be attributed to the way the experiment was set up. Sample and Bayes are always ready to bet, while Conf has to skip some of the trials if the ticket price falls within the bounds of its confidence interval. By imposing a limited number of tokens that could be used to place a bet, the playing field was somewhat levelled so that all players would place an approximately equal number of bets. (Conf might still place fewer bets if there were not enough "good" trials.)

We can consider setting other rules for the game. Suppose Conf is *forced* to bet? "Force" implies sanctions; sanctions imply value. The value involved in the sanction distorts the odds. The way to handle this is to allow Conf to refrain from buying and selling tickets, but to charge it a small penalty every time it exercises this option. This will increase the number of bets placed by Conf, while retaining the choice to decline to bet if necessary. The effects of imposing a penalty might be similar to that of decreasing the level of con-fidence $1 - \alpha$ in some cases, but it provides a more flexible way to manage the betting criterion.

We have shown that the confidence interval approach can outperform the subjective Bayesian approach in quite a broad range of circumstances. It should be noted that this is *not* a question involving differences of parameters, as it would be if one approach were superior when the actual relative frequency is close to 0, and the other were superior when the actual relative frequency is close to 0.5. The differences appear to be quite global. The results suggest that confidence methods, involving the acceptance or rejection of families of statistical hypotheses, can offer practical advantages over purely probabilistic methods. Not only do Bayesian methods require the assumption of a prior probability (often hard to justify), but there appears to be a practical difference in return between updating the "uninformative" prior by conditionalization, and updating (pruning) the set of possible distributions by the application of confidence methods at a certain level of confidence. This suggests that the role of acceptance is not that of providing a rough shorthand for probabilistic updating, but more important and fundamental.

Further work needs to be done to characterize more precisely the circumstances under which one approach is better than the other, by taking account of a more finely articulated Bayesian strategy. We should also be able to extend this method of performance measure to evaluate other approaches to uncertainty.

## Acknowledgement

This work was supported by the National Science Foundation and the Australian Research Council.

## References

[Carnap, 1950] Rudolf Carnap. *The Logical Foundations of Probability*, volume Chicago. University of Chicago Press, 1950.

[Cramér, 1951] Harald Cramér. *Mathematical Methods of Statistics*. Princeton University Press, Princeton, 1951.

[Dubois and Prade, 1992] Didier Dubois and Henri Prade. Belief change and possibility theory. In Peter Gardenfors, editor, *Belief Revision*, pages 142–182. Cambridge Univeristy Press, Cambridge, 1992.

[Mises, 1957] Richard Von Mises. *Probability Statistics and Truth*, volume London. George Allen and Unwin, 1957.

[Murtezaoglu, 1998] Bulent Murtezaoglu. PhD thesis, The University of Rochester, 1998.

[Russell, 1901] Bertrand Russell. Recent works on the principles of mathematics. *The International Monthly*, 4:83–101, 1901.

[Savage, 1954] L. J. Savage. *Foundations of Statistics*. John Wiley, New York, 1954.

[Shafer, 1976] Glenn Shafer. *A Mathematical Theory of Evidence*. Princeton University Press, Princeton, 1976.

[Walley, 1995] Peter Walley. Inferences from multinomial data: Learning about a bag of marbles. *Journal of the Royal Statistical Society*, B, 57, 1995.